\documentclass[a4paper,fleqn]{cas-dc}

\usepackage[numbers]{natbib}

\usepackage{algorithm}
\usepackage{algorithmic}
\usepackage{float}
\usepackage{subcaption}

\def\tsc#1{\csdef{#1}{\textsc{\lowercase{#1}}\xspace}}
\tsc{WGM}
\tsc{QE}
\tsc{EP}
\tsc{PMS}
\tsc{BEC}
\tsc{DE}

\begin{document}
\let\WriteBookmarks\relax
\def\floatpagepagefraction{1}
\def\textpagefraction{.001}

\shorttitle{GDEPO for Efficient RL}

\shortauthors{Zhengqing Yan et~al.}

\title [mode = title]{GDEPO: Group Dual-dynamic and Equal-right Advantage Policy Optimization for Sample-Efficient Reinforcement Learning}                   


%
\author[1]{Zhengqing Yan}[type=editor,
                        style=chinese,
                        orcid=0009-0001-2233-4242]

\author[2]{Fan Guo}[style=chinese]

\author[3]{Xinyang Liu}[style=chinese]

\author[1]{Yi Zhang}[style=chinese]

\author[3]{Yao Liu}[style=chinese]

\author[3]{Qi Liu}[style=chinese]

\author[3]{Jihao Huang}[style=chinese]

\author[1]{Kang Song}[style=chinese,]
\ead{songkangtju@tju.edu.cn}
\cormark[1]

\affiliation[1]{organization={Tianjin University, State Key Laboratory of Engines},
    addressline={No. 135 Yaguan Road, Jinnan District}, 
    city={Tianjin},
    postcode={300350}, 
    country={China}}

\affiliation[2]{organization={Hong Kong Centre for Logistics Robotics},
    addressline={Building 17W, Hong Kong Science Park},
    city={Hong Kong SAR},
    postcode={},
    country={China}}

\affiliation[3]{organization={Artificial Intelligence Center, Cylingo Group},
    addressline={No. Jia 48 Zhichun Road, Haidian District},
    city={Beijing},
    postcode={100080}, 
    country={China}}

\cortext[cor1]{Corresponding author}

\begin{abstract}
Automated Theorem Proving (ATP) represents a fundamental challenge in Artificial Intelligence (AI), requiring the construction of machine-verifiable proofs in formal languages such as Lean to evaluate AI reasoning capabilities. Reinforcement Learning (RL), particularly the high-performance Group Relative Policy Optimization (GRPO) algorithm, has emerged as a mainstream approach for this task. However, in ATP scenarios, GRPO faces two critical issues: when composite rewards are used, its relative advantage estimation may conflict with the binary feedback from the formal verifier; meanwhile, its static sampling strategy may discard entire batches of data if no valid proof is found, resulting in zero contribution to model updates and significant data waste. Repeated training on the same dataset causes the model to waste computational resources on already mastered problems. To address these limitations, we propose Group Dual-dynamic and Equal-right Advantage Policy Optimization (GDEPO), a method incorporating three core mechanisms: 1) dynamic additional sampling, which resamples ineffective batches until a valid proof is discovered; 2) equal-right advantage, decoupling the sign of the advantage function (based on correctness) from its magnitude (modulated by auxiliary rewards) to ensure stable and correct policy updates. Meanwhile, it increases additional effective computation that cannot be increased by traditional methods to improve model performance; and 3) dynamic additional iterations, applying extra gradient steps to initially failed but eventually successful samples to accelerate learning on challenging cases. Experiments conducted on three datasets of varying difficulty (MinF2F-test, MathOlympiadBench, PutnamBench) confirm the effectiveness of GDEPO, while ablation studies validate the necessity of its synergistic components. The proposed method enhances data utilization and optimization efficiency, offering a novel training paradigm for ATP.
\end{abstract}


\begin{highlights}
\item Resamples ineffective GRPO rollouts to maximize data utility.
\item Decouples advantage sign from auxiliary reward magnitude in GRPO.
\item Adds adaptive extra iterations for challenging samples.
\end{highlights}


\maketitle

\section{Introduction}
The rapid advancement of Large Language Models (LLMs) has opened new avenues for applying Artificial Intelligence (AI) to formal reasoning. Automated Theorem Proving (ATP), a core challenge in AI, requires systems to construct machine-verifiable mathematical proofs in formal languages such as Lean\cite{de2015lean, moura2021lean4, ying2024leanworkbook}. Recent progress in ATP on complex theorem-proving tasks stems from the strong prior knowledge and generative capabilities of LLMs, coupled with the availability of extensive training and validation datasets\cite{zheng2022miniff, lin2025goedelv2, tsoukalas2024putnambench, gao2025herald, hu2025minictx, wang2024proving}. Notably, AlphaGeometry\cite{trinh2024solving} has demonstrated problem-solving performance approaching International Mathematical Olympiad (IMO) medalist levels, signaling a transition of LLM-driven ATP from theoretical exploration toward practical deployment.

A prevailing approach integrates LLMs with Reinforcement Learning (RL) algorithms, leveraging precise feedback from formal verifiers to guide policy optimization. Among these, Group Relative Policy Optimization (GRPO)\cite{shao2024grpo} has gained widespread adoption due to its strong empirical performance. However, GRPO faces three critical limitations in the ATP setting. First, when composite rewards are used, its advantage computation may assign negative advantages to correct trajectories due to interference from other reward components—a mathematical possibility formally detailed in Section VI. To avoid this issue, many works resort to a single binary correctness reward, which forfeits RL’s essential role in aligning LLM outputs with human preferences. Second, data efficiency is poor: under a single binary reward, groups containing only incorrect trajectories yield zero advantages across all samples, leading to the discarding of entire batches despite potentially informative partial reasoning—particularly problematic when challenging samples dominate the training set. Furthermore, repeated training on the same dataset wastes computational resources on already mastered samples and increases the risk of overfitting. Third, a fixed number of optimization iterations treats all samples uniformly, ignoring their varying learning potential.

\begin{figure*}[!t]
\centering
\includegraphics[width=7in]{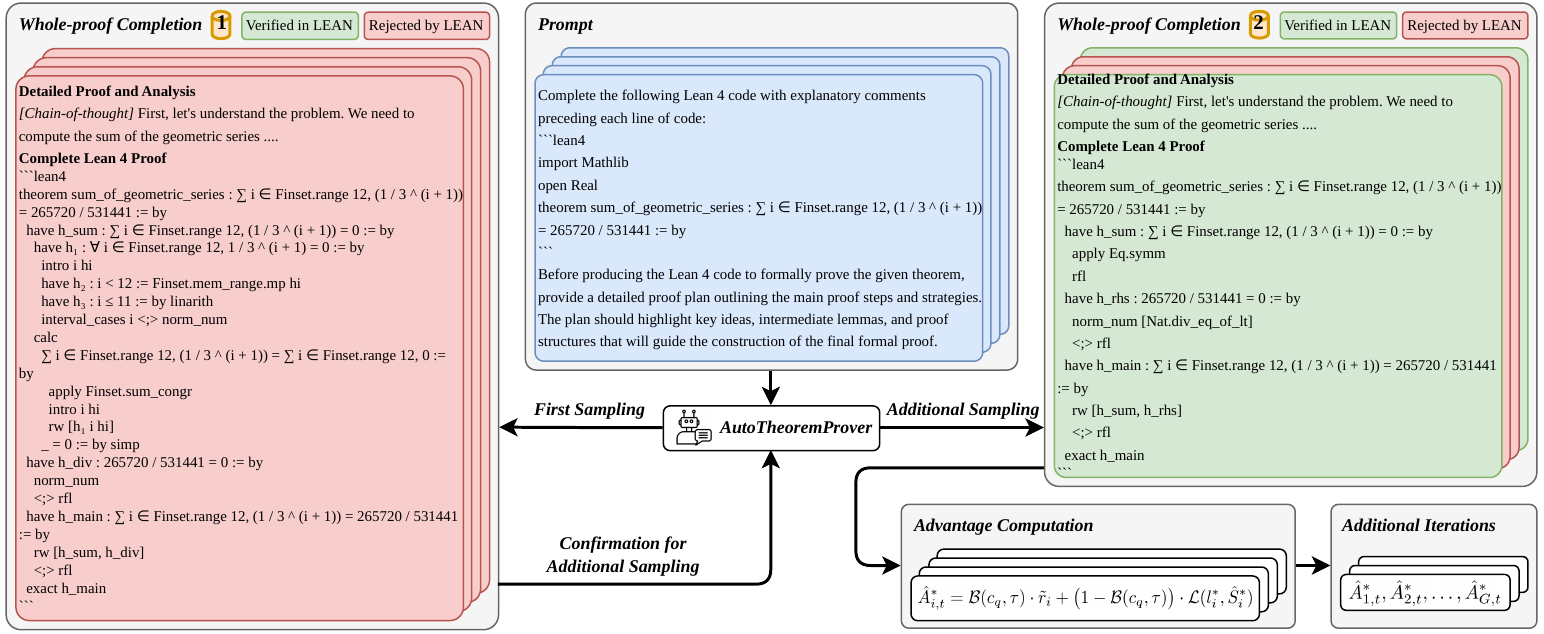}
\centering
\caption{The detailed training procedure of GDEPO: LLMs serve as auto-theorem provers by receiving a standardized prompt containing $G$ identical problem instances, generating whole-proof completions, and verifying them through a Lean4 server\cite{santos2025kimina}. A sample is deemed ineffective if all outputs in the group are incorrect. For the same prompt, sampling continues up to a maximum number of attempts until at least one correct proof is obtained. When the within-group accuracy falls below a predefined threshold, the correctness feedback from the verifier contributes a sign term to the advantage function, while other signals determine its magnitude. For samples initially classified as ineffective but later yielding valid proofs through subsequent sampling, multiple extra gradient update steps} are performed to refine the model parameters.
\label{fig:output_show}
\end{figure*}

To address the aforementioned limitations of GRPO when applied to formal theorem proving, we propose Group Dual-dynamic and Equal-right Advantage Policy Optimization (GDEPO), a novel training framework whose workflow is illustrated in Fig.\ref{fig:output_show}. Our main contributions are summarized as follows:

\subsection{Dynamic additional sampling for maximizing data utility}
Targeting the whole-batch data waste caused by GRPO’s static single-round sampling strategy, we design a dynamic resampling mechanism. For training queries where all initial sampled outputs are invalid, the system performs additional sampling rounds until at least one correct proof is obtained or the maximum attempt limit is reached. This mechanism greatly improves the exploration efficiency for low-probability correct solutions on challenging problems, and injects effective extra computation into the model that cannot be achieved by traditional static training paradigms. This module is detailed in Section III-A, and its performance gain on challenging samples is verified in Section V-B.
\subsection{Equal-right advantage for theoretically sound optimization under composite rewards}
To resolve the fundamental conflict that correct trajectories may receive negative advantages when using composite rewards in standard GRPO, we propose a decoupled advantage computation paradigm. The sign of the advantage function (which determines the direction of policy update) is governed exclusively by the binary correctness feedback from the Lean verifier, while auxiliary rewards only modulate the magnitude of the advantage (which controls the strength of policy update). This mechanism guarantees correct optimization directions under arbitrary reward weight combinations, while enabling multi-objective reward shaping for better proof quality. This module is elaborated in Section III-B, and its effectiveness in balancing correctness and proof conciseness is validated in Section V-C.
\subsection{Dynamic additional iterations for accelerated mastery of challenging problems}
Aiming at the limitation that fixed optimization iterations ignore the heterogeneous learning value of different samples, we propose an adaptive iteration allocation strategy. For medium-challenging samples that fail initial sampling but yield valid proofs after resampling, we apply multiple extra gradient update steps to concentrate computational resources on high-value learning trajectories. This module is presented in Section III-C, and its synergistic effect with dynamic sampling is confirmed by ablation studies in Section V-D.
\subsection{Systematic empirical validation across multi-difficulty benchmarks}
We conduct comprehensive experiments on three benchmarks with progressively increasing difficulty: MiniF2F-test, MathOlympiadBench, and PutnamBench. Results demonstrate that GDEPO consistently outperforms state-of-the-art baselines, with a notable 84\% performance improvement on the most challenging PutnamBench. Ablation studies further verify the necessity of each component and their synergistic interaction, providing a novel training paradigm for reinforcement learning in formal reasoning tasks.

GDEPO can be viewed as an adaptive allocation of exploration and optimization effort: additional computation is concentrated on challenging yet learnable problems, while the verifier preserves the correct direction of policy improvement.

\section{Related Work}

\subsection{Macroscopic Paradigm for Formal Theorem Proving}

Existing algorithms aim to enhance proof success rates by innovating the inference paradigm.

The end-to-end generation paradigm trains models to directly output complete proof sequences that can be verified by formal verifiers such as Lean\cite{de2015lean, moura2021lean4, ying2024leanworkbook}. Representative works include DeepSeek-Prover\cite{xin2024deepseekprover}, Goedel-Prover\cite{lin2025goedel}, and Kimina-Prover\cite{wang2025kimina}. This paradigm leverages LLMs' sequence generation capabilities, condensing the proof process into a single forward generation. It offers advantages of simple architecture and efficient reasoning. However, its performance is significantly constrained by the reliability of single-generation outputs, posing robustness challenges for complex theorems.

The proof search--based paradigm guides the model to incrementally construct proofs by exploring derivation paths at each step using feedback from a verifier. For instance, systems such as DeepSeek-Prover v1.5\cite{xin2024deepseekprover1.5}, BFS-Prover\cite{xin2025bfs}, and ABEL\cite{gloeckle2024abel} employ tree search algorithms—including Monte Carlo tree search or breadth-first search—to explore multiple proof trajectories and iteratively assemble valid proofs. This approach substantially enhances the robustness and provable scope of the system, albeit at the cost of considerable computational and time overhead.

To integrate the strengths of these paradigms or mitigate their inherent bottlenecks, a series of hybrid methods and auxiliary strategies have been proposed. These include leveraging retrieval-augmented generation (RAG) to introduce relevant theorems\cite{yang2023leandojo}; using informal proof sketches to guide formal step generation\cite{jiang2023draft, yang2023leandojo}; and designing self-verification and iterative optimization loops\cite{ji2025leanabell}. The "verifier-in-the-loop" interactive strategy\cite{ren2025deepseekproverv2, rajaee2025local} further highlights the value of iterative improvement through refined feedback. Additionally, hybrid architectures like DSP+\cite{cao2025reviving} and Delta-Prover\cite{zhou2025solving} adopt a two-stage method of "first generating a framework, then filling in details," although their effectiveness typically relies on costly ultra-large-scale foundational models, such as DSP+ using the 671B parameter DeepSeek-R1\cite{guo2025deepseek} or DeepSeek-V3\cite{liu2024deepseekv3}, while Delta-Prover utilizes Gemini\cite{comanici2025gemini}.

Regardless of innovations in macro system architecture, ultimate performance remains limited by the underlying strategy's ability to learn from formalized feedback. The efficiency, stability, and utilization of feedback signals by the algorithm directly determine whether the system can fully exploit data value and achieve reliable optimization.

\subsection{Inadequate Adaptability of Policy Optimization Algorithms in Formal Theorem Proving}

Inspired by informal LLMs in natural language processing, recent foundational work in formal theorem proving has adopted similar RL-based policy optimization strategies. Current formal theorem provers primarily rely on policy optimization algorithms originally developed for informal LLM training, such as Proximal Policy Optimization (PPO)\cite{schulman2017proximal}, Direct Preference Optimization (DPO)\cite{rafailov2023dpo}, and Group Relative Policy Optimization (GRPO)\cite{shao2024grpo}, which is widely employed by state-of-the-art (SOTA) systems\cite{ren2025deepseekproverv2, wang2025kimina, lin2025goedelv2}. While these algorithms have demonstrated effectiveness in their native domains—such as language alignment and complex reasoning—they exhibit significant adaptation limitations when transferred to the formal theorem proving setting, where feedback is precise and binary.

\subsubsection{Limitations of the PPO and Reward Modeling Paradigm}

PPO and its associated reward-model-based paradigm rely on the assumption that the reward function can accurately and stably capture complex, composite qualities—such as proof correctness and conciseness. In formal theorem proving, however, designing or learning such a reward model is highly challenging, often leading to training instability and reward overfitting, which in turn hinders effective multi-objective optimization.

\subsubsection{Limitations of the Direct Preference Optimization and Preference Optimization Paradigm}

DPO enhances training simplicity by directly leveraging preference data, thereby bypassing explicit reward model training. However, its optimization signal relies on global pairwise rankings, which are ill-suited to incorporate the highly structured, fine-grained feedback inherent to formal environments. Consequently, DPO struggles to support localized, incremental refinement of complex proof processes, limiting its effectiveness in interactive theorem proving scenarios that require stepwise correction.

\subsubsection{Limitations of the Group Relative Policy Optimization Paradigm}

GRPO has become the core optimization algorithm in current SOTA provers—such as DeepSeek-Prover-V2\cite{ren2025deepseekproverv2}, Kimina-Prover\cite{wang2025kimina}, and Goedel-Prover-V2\cite{lin2025goedelv2}—due to its elimination of a separate reward model and computational efficiency. However, its fundamental mechanism, which relies on relative comparisons among samples within a group, inherently conflicts with the absolute correctness criterion demanded by formal verification. This tension manifests in two critical issues. First, when composite rewards are employed, GRPO’s advantage computation may assign negative advantages to otherwise correct proof trajectories due to suboptimal scores in other reward components—a mathematical possibility rigorously established in SectionVI. To avoid this, practitioners often resort to binary rewards alone, thereby forfeiting the core capability of RL to align large language model outputs with nuanced human preferences, such as proof conciseness or elegance. Second, GRPO exhibits low data utilization efficiency: in batches containing a high proportion of challenging samples, using a single binary reward frequently results in zero advantages for all trajectories within a group. Consequently, all sampled attempts that fail to produce a valid proof are collectively discarded, wasting valuable training data. To mitigate these risks, existing SOTA systems\cite{ren2025deepseekproverv2, wang2025kimina, lin2025goedelv2} adopt a conservative strategy—using only binary rewards and deliberately forgoing the potential benefits of composite rewards. Moreover, data batches comprising exclusively correct or incorrect samples are often discarded outright\cite{yu2025dapo}. These compromises yield dual adverse consequences: (i) the precise and rich feedback from formal verifiers remains underutilized, and (ii) the already scarce high-quality training data suffers from further reduced effective usage, impeding the model’s ability to learn from challenging examples. Additionally, the use of a fixed number of optimization iterations treats all samples uniformly, ignoring inherent differences in their learning value. As a result, the current training framework leaves considerable room for more efficient exploitation of formal proof data.

In summary, the inherent incompatibility between existing policy optimization algorithms—particularly GRPO, which serves as the de facto standard—and the formal verification environment constitutes a fundamental bottleneck that impedes efficient data utilization, reliable multi-objective optimization, and effective learning from challenging examples. Developing a novel optimization framework that retains computational efficiency while fundamentally resolving these conflicts has thus become pivotal to advancing the field.

\begin{figure}
\centering
\includegraphics[width=0.95\linewidth]{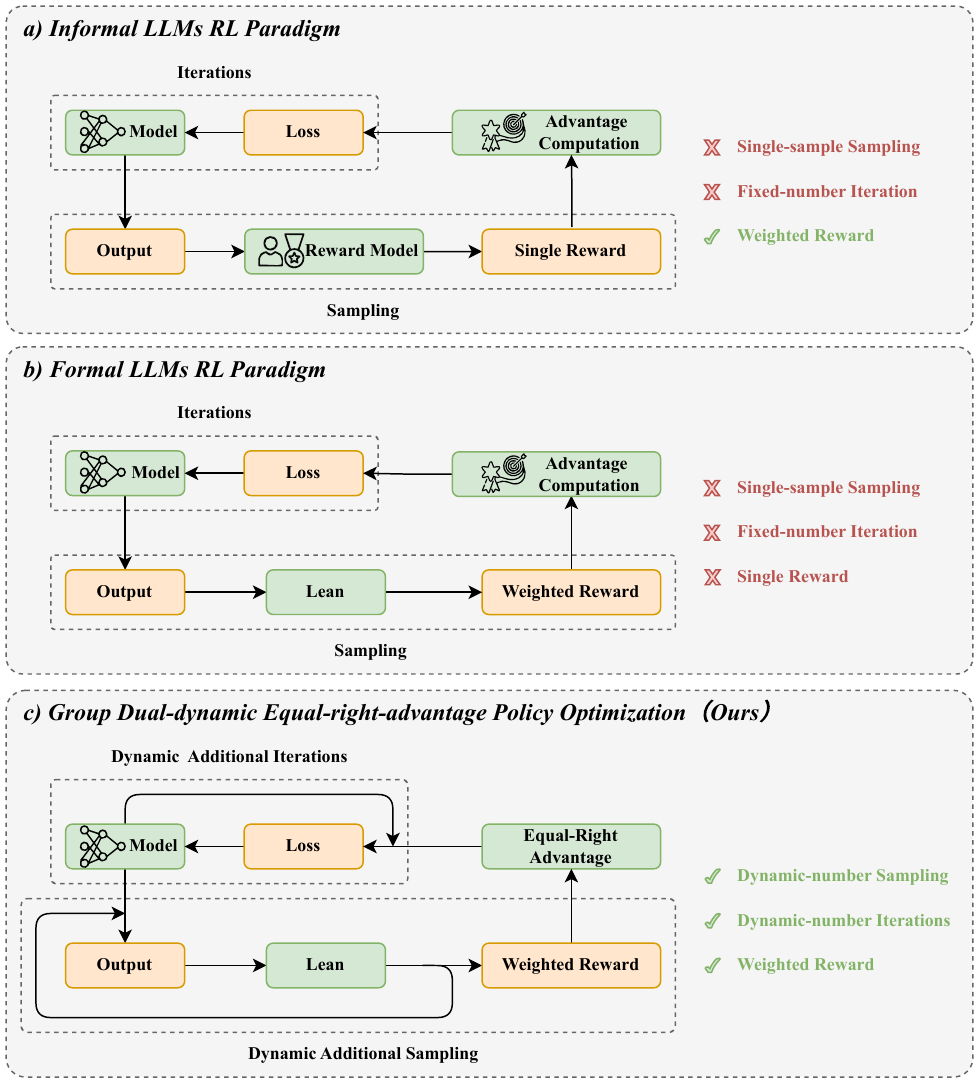}
\caption{Overall architecture of the GDEPO framework and its comparison with conventional paradigms.}
\label{fig:relatedWork}
\end{figure}

The overall architecture of the GDEPO framework and its comparison with conventional paradigms are illustrated in Fig.\ref{fig:relatedWork}. The informal large language model RL paradigm, depicted in Fig.\ref{fig:relatedWork}a, relies on a trained reward model to generate scalar rewards. However, such rewards lack the precise binary signal required to guarantee formal correctness. In contrast, the standard formal ATP paradigm, shown in Fig.\ref{fig:relatedWork}b, employs a Lean verifier as the ground-truth judge but inherits two rigid design choices from generic RL: (i) a single-round sampling mechanism that struggles to explore low-probability correct solutions for complex problems, and (ii) a fixed number of iterations applied uniformly across all training samples, disregarding their varying learning utility. Moreover, the advantage estimation in this paradigm faces a dilemma: using a composite reward function often leads to misaligned gradient directions, whereas relying solely on the binary correctness reward results in poor data efficiency and weakens the ability of RL optimization to reflect desired output preferences.

As shown in Fig.\ref{fig:relatedWork}c, the GDEPO algorithm builds upon the GRPO framework with several key enhancements. During training, the framework identifies ineffective samples and performs additional rollouts until valid ones are obtained, thereby maximizing data utilization. It then computes advantages using a redesigned function that enables proper integration of composite rewards for policy optimization. Finally, the number of policy update iterations is dynamically adjusted based on the difficulty of each sample, accelerating learning on challenging instances.

\section{Group Dual-dynamic and Equal-Right Advantage Policy Optimization}

GDEPO was proposed to address the aforementioned limitations of standard RL in formal theorem proving. The framework simultaneously improves data efficiency and optimization rationality.

\begin{figure*}[!t]
\centering
\includegraphics[width=0.95\textwidth]{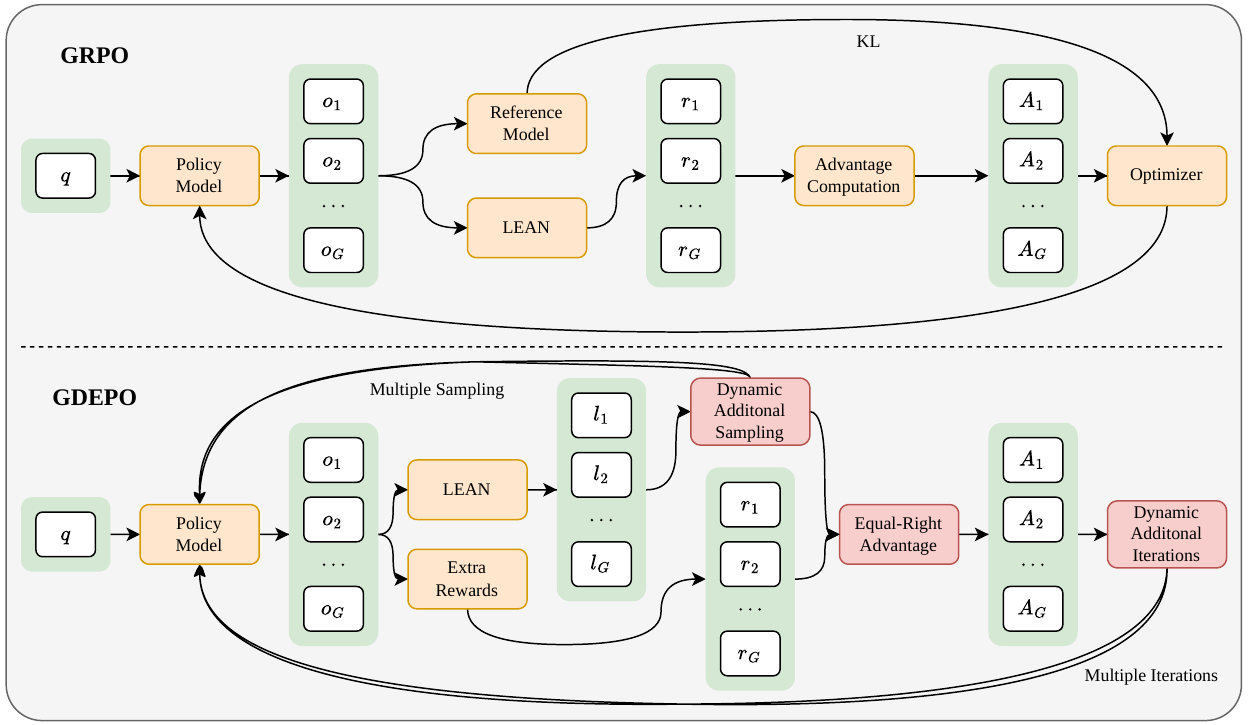}
\caption{Overall training framework of GDEPO in comparison with GRPO.}
\label{fig:mathod}
\end{figure*}

As illustrated in Fig.\ref{fig:mathod}, GDEPO overcomes these design constraints through two dynamic modules and a novel advantage computation method. The first module, \textit{dynamic additional sampling}, actively increases the number of rollouts until a valid proof is found, thereby maximizing the utility of each training instance. The second module, \textit{equal-right advantage}, decouples the verifier’s authoritative binary correctness signal from auxiliary quality-based rewards, ensuring that policy updates follow a theoretically sound optimization direction. The third component, \textit{dynamic additional iterations}, allocates extra policy update steps based on the empirical difficulty of each sample. Together, these components form an integrated adaptive learning loop that concentrates exploration and computational resources on the most informative and challenging proof trajectories, leading to robust policy improvement and efficient data utilization.

The implementation details of these three core modules are elaborated below.

\subsection{Dynamic Additional Sampling}

Standard GRPO performs only a single round of sampling per query. When handling challenging samples, the model may assign low probabilities to correct solutions\cite{he2025rewarding}. Under this fixed single-round scheme, if all sampled outputs are incorrect, the resulting batch contains no positive learning signal. Discarding such batches not only wastes computational resources but also overlooks potentially valid candidates with low generation probabilities. To mitigate this issue, dynamic additional sampling is introduced to enhance the exploration of low-probability correct solutions.

The module follows a simple yet efficient design: if the initial sampling for a query yields no correct output, the system performs additional sampling on the same query until either a valid output is obtained or a preset maximum number of attempts is reached. The detailed pipeline is summarized in Algorithm\ref{alg:Dynamic Additional Sampling}. For a given query $q$, the set of outputs from the $k$-th sampling round is denoted as follows:
\begin{equation}
\label{eq:sampling}
\mathcal{G}_q^{(k)} = \left\{ o_1^{(k)}, o_2^{(k)}, \ldots, o_G^{(k)} \right\},
\end{equation}
where $G$ denotes the fixed number of outputs per group. This output group corresponds to a core binary reward set
\begin{equation}
\label{eq:core_rewards}
\mathcal{L}_q^{(k)} = \left\{ l_1^{(k)}, l_2^{(k)}, \ldots, l_G^{(k)} \right\},
\end{equation}
with each $l_i^{(k)} \in \{1, -1\}$, where $1$ indicates a correct output and $-1$ an incorrect one (i.e., the Lean4 reward), and an auxiliary reward set
\begin{equation}
\label{eq:aux_rewards}
\mathcal{R}_q^{(k)} = \left\{ \mathbf{r}_1^{(k)}, \mathbf{r}_2^{(k)}, \ldots, \mathbf{r}_G^{(k)} \right\},
\end{equation}
where each $\mathbf{r}_i^{(k)} = \left( r_{i,1}^{(k)}, r_{i,2}^{(k)}, \ldots, r_{i,K}^{(k)} \right)$ is a $K$-dimensional auxiliary reward vector with components satisfying $r_{i,j}^{(k)} \in [0, 1]$ for all $j = 1, \ldots, K$.

Correctness is our primary criterion; we aim to ensure that among the $G$ trajectories generated for query $q$, at least one is correct to facilitate effective model optimization. Thus, the dynamic sampling terminates when either a correct output exists in the group, i.e., $\exists l_i^{(k)} = 1$, or the number of sampling iterations reaches the maximum limit $n$. The final valid output group for query $q$ is denoted as $\mathcal{G}_q^*$, along with its associated valid rewards $\mathcal{L}_q^*$ and $\mathcal{R}_q^*$. Formally, this is expressed as

\begin{subequations}
    \label{eq:final_group}
    \begin{align}
        \mathcal{G}_q^* &= \mathcal{G}_q^{(k^*)}, \\
        \mathcal{L}_q^* &= \mathcal{L}_q^{(k^*)}, \\
        \mathcal{R}_q^* &= \mathcal{R}_q^{(k^*)}.
    \end{align}
\end{subequations}
where,$k^* = \min\{ k \mid k = n \text{ or } \exists\, l_i^{(k)} = 1, k \in [0,n],  k \in \mathbb{Z} \}$; $q \sim P(Q)$ denotes a single problem sampled from the distribution of mathematical theorem-proving tasks; $l_i^{(k)} \in \{1, -1\}$ indicates the binary correctness label (with $1$ denoting correct and $-1$ denoting incorrect) associated with the $i$-th trajectory in the $k$-th sampling; $n \in \mathbb{Z}$ is the user-specified maximum number of sampling (e.g., $n = 3$).

RL training is typically terminated when the reward ceases to increase. Once the reward has converged, additional computation yields little or no further gain in model performance. The dynamic additional sampling module addresses this limitation by increasing the model's exploration of complex problems without manual intervention, while avoiding the inefficiency of repeated training on the same dataset, which otherwise wastes computation on problems the model has already mastered.

\begin{algorithm}[ht]
\caption{Dynamic Additional Sampling.}\label{alg:Dynamic Additional Sampling}
\begin{algorithmic}[1]
\REQUIRE maximum number of sampling $n \in \mathbb{Z}$, the query $q$ \\
\ENSURE the number of sampling $k$, valid trajectory group $\mathcal{G}_q^*$, core reward $\mathcal{L}_q^*$, auxiliary reward $\mathcal{R}_q^*$ \\
\STATE \hspace{0cm}initialize $k \leftarrow 0$
\STATE \hspace{0cm}\textbf{for } $k = 0, 1, \dots, n$ \textbf{ do}
\STATE \hspace{0.5cm}do sampling by \textbf{Eq.\ref{eq:sampling}-\ref{eq:aux_rewards}} to get $\mathcal{G}_q^{(k)}, \mathcal{L}_q^{(k)}, \mathcal{R}_q^{(k)}$
\STATE \hspace{0.5cm}update the counter $k \leftarrow k + 1$
\STATE \hspace{0.5cm}\textbf{if } $\exists l_i^{(k)} = 1$ \textbf{ then}
\STATE \hspace{1.0cm}$\text{break}$
\STATE \hspace{0.5cm}\textbf{end}
\STATE \hspace{0cm}\textbf{end}
\STATE \hspace{0cm}do store the latest valid value by \textbf{Eq.\ref{eq:final_group}} to get $\mathcal{G}_q^*, \mathcal{L}_q^*, \mathcal{R}_q^*$
\STATE \hspace{0cm}\textbf{return} $k,\mathcal{G}_q^*, \mathcal{L}_q^*, \mathcal{R}_q^*$ 
\end{algorithmic}
\end{algorithm}

\subsection{Equal-Right Advantage}

The \textit{equal-right advantage} is introduced to resolve a fundamental conflict that arises when applying GRPO to formal theorem proving tasks. Standard GRPO computes advantage values through intra-group comparisons. As shown in SectionVI, when composite rewards are used, it is mathematically possible for a trajectory verified as correct by Lean to receive a negative advantage due to unfavorable auxiliary rewards. This can cause the optimizer to suppress already-validated correct solutions, leading to training instability.

The detailed code pipeline is summarized in Algorithm\ref{alg:Equal-Right Advantage}. Equal-right advantage decouples the supervision signal: the sign of the advantage—i.e., the direction of policy update—is determined solely by the binary label from the verifier, while auxiliary rewards modulate only the magnitude of the advantage—i.e., the update strength. This mechanism guarantees that all correct proofs receive positive signals and all incorrect ones receive negative signals, preserving the authority of verification outcomes while enabling multi-objective reward shaping. Importantly, it yields correct policy update directions even when all $G$ trajectories for a query are either entirely correct or entirely incorrect.

Furthermore, the module adapts to sample difficulty: when the passing rate of sampled trajectories is low, equal-right advantage is activated to prioritize proof correctness; once the passing rate exceeds a preset threshold $\tau$, the algorithm reverts to the standard GRPO advantage computation, allowing auxiliary rewards to refine proof quality within a reliably correct solution set. The equal-right advantage is computed as follows:

\begin{subequations}
    \label{eq:aux_A_hat}
    \begin{align}
        \hat{A}_{i,t}^* &= 
        \begin{cases} 
            \tilde{r}_i, & c_q > \tau , \\ 
            l_i^* \cdot \mathcal{L}(l_i^*, \hat{S}_i^*), & c_q \leq \tau ,
        \end{cases} \\
        \mathcal{L}(l_i^*, \hat{S}_i^*) &= \frac{l_i^* + 1}{2} \cdot \hat{S}_i^* + \frac{1 - l_i^*}{2} \cdot (1 - \hat{S}_i^*), \\
        \hat{S}_i^* &= 
        \begin{cases} 
            0.5 & \text{if } S_{\text{max}}^* = S_{\text{min}}^*, \\
            \dfrac{S_i^* - S_{\text{min}}^*}{S_{\text{max}}^* - S_{\text{min}}^*} & \text{otherwise}.
        \end{cases}
    \end{align}
\end{subequations}
Here, $t$ denotes the token index, and $\hat{A}_{i,t}^*$ remains constant across all tokens within the same trajectory; $c_q \in [0,1]$ denotes the pass rate of the query $q$; $\tau \in (0,1)$ is a predefined threshold on pass rate; $\tilde{r}_i = \frac{S_i^* - \text{mean}(S^*)}{\text{std}(S^*)}$ represents the normalized advantage of the $i$-th trajectory as used in conventional GRPO, where $S_i^* = \sum_{k=1}^K w_k \cdot r_{i,k}^*$ is the weighted sum of auxiliary rewards for trajectory $i$, $\text{mean}(S^*)$ and $\text{std}(S^*)$ are the mean and standard deviation of $\{S_j^*\}_{j=1}^G$ over the valid output group, respectively; $l_i^* \in \{1, -1\}$ is the core binary correctness label for the $i$-th trajectory in the valid group ($1$ for correct, $-1$ for incorrect); $r_{i,k}^* \in [0, +\infty)$ is the $k$-th non-negative auxiliary reward associated with trajectory $i$ ($k \geq 1$); $w = (w_1, w_2, \dots, w_K)$ is a weight vector for auxiliary rewards satisfying $w_k \geq 0$ and $\sum_{k=1}^K w_k = 1$; $S_{\text{max}}^* = \max_{j=1,\dots,G} S_j^*$ and $S_{\text{min}}^* = \min_{j=1,\dots,G} S_j^*$ denote the maximum and minimum of the auxiliary reward sums within the valid group; and $\hat{S}_i^*$ is the min–max normalized value of $S_i^*$ mapped into $[0,1]$.

Finally, $T_i$ denotes the token length of the $i$-th trajectory, with all tokens in the trajectory sharing the same advantage value $\hat{A}_{i,t}$.
$\mathcal{G}_q^*, \mathcal{L}_q^*, \mathcal{R}_q^*$

\begin{algorithm}[ht]
\caption{Equal-Right Advantage.}\label{alg:Equal-Right Advantage}
\begin{algorithmic}[1]
\REQUIRE valid trajectory group $\mathcal{G}_q^{(*)}$, core binary reward $\mathcal{L}_q^*$, auxiliary reward $\mathcal{R}_q^*$, weights of auxiliary rewards $w = (w_1,\dots,w_K)$, the pass rate of the query $c_q$, accuracy threshold $\tau$
\ENSURE advantage value list of the query $\hat{\mathcal{A}}_{\text{list}}^*$
\STATE do weighted auxiliary rewards computation to get \\ $S^*_{\text{all}} \leftarrow \{ \sum_{k=1}^K w_k \cdot r_{i,k} \mid i=1,\dots,G \}$
\STATE \textbf{for } $i = 1$ \textbf{to } $G$ \textbf{ do}
\STATE \hspace{0.5cm} do advantage value per token computation by \\ \text{ } \text{ } \text{ } \textbf{Eq.\ref{eq:aux_A_hat}} to get $\hat{A}_{i,t}$
\STATE \hspace{0.5cm} do advantage values for each token aggregate to get \\ \text{ } \text{ } \text{ } $\hat{\mathcal{A}}_{\text{list}}^*.append(\hat{A}_{i,t}^*)$
\STATE \textbf{end}
\STATE \textbf{return} $\hat{\mathcal{A}}_{\text{list}}^*$
\end{algorithmic}
\end{algorithm}

Thus, correctness determines whether a trajectory should be promoted or suppressed, whereas auxiliary preferences determine how strongly it should be updated.

The component is relatively critical in the method. The equal-right advantage helps to utilize the training data that would otherwise be discarded while maintaining a reasonable model update direction, thereby increasing the effective amount of computation without incurring additional sampling overhead, and potentially improving model performance.

\subsection{Dynamic Additional Iterations}

Although the dynamic additional sampling module enhances the exploration of low-probability correct solutions, it introduces a new issue of sample heterogeneity: proofs obtained directly in the initial sampling and those discovered only after repeated exploration possess markedly different instructional values for model optimization. The latter—requiring sustained search to succeed—deserve deeper RL attention. Applying identical single-step gradient updates to both cases fails to exploit the richer learning potential of the latter.

To address this, \textit{dynamic additional iterations} are proposed to allocate computational resources proportionally to each sample’s learning value. The detailed code pipeline is summarized in Algorithm\ref{alg:Dynamic Additional Iterations}. The principle is straightforward: easy queries (correct on the first attempt) and persistently unsolvable hard queries undergo only one update iteration, whereas medium-challenging queries—those that yield a correct proof only after multiple sampling rounds—are subjected to $m$ iterations to accelerate optimization. Formally, for a query $q$, if its valid output set $\mathcal{G}_q^*$ contains at least one correct trajectory and was not produced in the first sampling round, then all $G$ trajectories for $q$ are used in $m$ consecutive policy updates. The overall objective function is expressed as:

This design adaptively concentrates computational effort on parameter updates with higher pedagogical value—particularly those arising from successful multi-round searches. Consequently, the model more rapidly internalizes knowledge for solving theorems that are challenging yet solvable.
\begin{equation}
    \label{eq:J_final_compact}
    \mathcal{J}(\theta) = \mathbb{E}_{\substack{q \sim P(Q) \\ \mathcal{G}_q^* \sim \pi_{\theta_{\text{old}}^{(s)}}}}\!\left[\frac{1}{G T_i}\sum_{\substack{i=1 \\ t=1}}^{G,\,T_i}\min \bigl[\rho_{i,t}^{(s)}\hat{A}_{i,t}^*,\bar{\rho}_{i,t}^{(s)}\hat{A}_{i,t}^*\bigr]\right]
\end{equation}
where the auxiliary operators are defined as
\begin{subequations}
    \label{eq:aux_J_final_compact}
    \begin{align}
        \rho_{i,t}^{(s)} &= \frac{\pi_{\theta^{(s)}}(o_{i,t}^* \mid q, o_{i,<t}^*)}{\pi_{\theta_{\text{old}}^{(s)}}(o_{i,t}^* \mid q, o_{i,<t}^*)}, \\
        \bar{\rho}_{i,t}^{(s)} &= \text{clip}\bigl(\rho_{i,t}^{(s)},\, 1 - \varepsilon_{\text{low}},\, 1 + \varepsilon_{\text{high}}\bigr).
    \end{align}
\end{subequations}
Here, $\rho_{i,t}^{(s)}$ denotes the policy ratio for the $t$-th token of the $i$-th trajectory during the $s$-th iteration; $\mathcal{C}(x) = \text{clip}(x, 1 - \varepsilon_{\text{low}}, 1 + \varepsilon_{\text{high}})$ is an asymmetric clipping function with user-defined bounds $\varepsilon_{\text{low}}, \varepsilon_{\text{high}} > 0$.

\begin{algorithm}[ht]
\caption{Dynamic Additional Iterations.}\label{alg:Dynamic Additional Iterations}
\begin{algorithmic}[1]
\REQUIRE advantage value list of the query $\hat{\mathcal{A}}_{\text{list}}$, the number of sampling $k$, the number of iterations $m$, upper and lower bounds on the clipping ratio $\varepsilon_{high}$, lower bounds on the clipping ratio $\varepsilon_{low}$, core binary reward $\mathcal{L}_q^*$
\ENSURE updated policy $\pi_\theta$
\STATE \textbf{if } $\exists l_i^{(*)} = 1$ \textbf{and} $k > 1$ \textbf{ then}
\STATE \hspace{0.5cm} \textbf{for } $\text{iter} = 0, 1, \dots, m$ \textbf{ do}
\STATE \hspace{1.0cm} do policy updating with loss by \textbf{Eq.\ref{eq:J_final_compact}-\ref{eq:aux_J_final_compact}} to get $\pi_\theta$ \\
\STATE \textbf{else}
\STATE \hspace{0.5cm} do policy updating with loss by \textbf{Eq.\ref{eq:J_final_compact}-\ref{eq:aux_J_final_compact}} to get $\pi_\theta$
\STATE \textbf{return} $\pi_\theta$
\end{algorithmic}
\label{alg2c}
\end{algorithm}

\section{DATASET AND TRAINING}

\subsection{Dataset for Training}

The experiments are conducted on the publicly available Lean4\cite{moura2021lean4}  formalized mathematical theorem dataset. To rigorously evaluate GDEPO's core capability—learning from challenging samples, we construct a targeted high-difficulty training set. The data are sourced from FineLeanCorpus\cite{peng2025criticlean}, a high-quality dataset comprising formally verified propositions with broad domain coverage and inherently high difficulty. To further isolate highly challenging instances, we use the pass@32 performance of the DeepSeek-Prover-V2-671B model as a filtering criterion, retaining only those samples for which no correct proof is found across 32 independent sampling attempts.

\subsection{Benchmarks}

We employ three widely adopted benchmarks for evaluating the mathematical theorem-proving performance of LLMs, ordered by increasing difficulty. 

\subsubsection{MiniF2F-test}

MiniF2F-test, the test split of MiniF2F\cite{zheng2022miniff}, consists of 244 Lean-formalized problems derived from high school mathematical olympiads, including the American Mathematics Competitions (AMC), the American Invitational Mathematics Examination (AIME), and the International Mathematical Olympiad (IMO). In this study, we adopt the cleaned version of MiniF2F provided by the Kimina team\cite{wang2025kimina}, which ensures reliability through 13-gram decontamination, removal of duplicates, and correction of unsolvable instances.

\subsubsection{MathOlympiadBench}

MathOlympiadBench\cite{lin2025goedelv2} raises the difficulty further by comprising 360 manually verified formal problems, each associated with a unique, complete, and Mathlib-compatible theorem along with an informal statement. The problems are drawn from the Compfiles database and the IMOSLLean4 codebase, forming a rigorously curated set of medium-to-high-difficulty proof tasks.

\subsubsection{PutnamBench}

PutnamBench\cite{tsoukalas2024putnambench} represents the most challenging evaluation in our study, featuring 644 university-level problems from the William Lowell Putnam Mathematical Competition held between 1962 and 2023. Covering algebra, analysis, number theory, geometry, combinatorics, probability, and set theory, this benchmark demands advanced mathematical insight and precise formal reasoning.

The aforementioned benchmarks collectively form a rigorous multi-level evaluation framework for assessing the model's generalization capability across continuously varying levels of mathematical complexity.

\subsection{Baselines}

We compare GDEPO-Prover with three SOTA models representing leading paradigms in the field of ATP, ensuring a diverse evaluation across model architectures and optimization strategies.

\subsubsection{DeepSeek-Prover-V2}

DeepSeek-Prover-V2\cite{ren2025deepseekproverv2} adopts a recursive subgoal decomposition paradigm. Its training process integrates multiple stages: problem decomposition and formalization using DeepSeek-V3; recursive subgoal solving with a 7 billion parameter model; curriculum learning for task generation; RL via GRPO algorithm, complemented by supervised fine-tuning and knowledge distillation.

\subsubsection{Kimina-Prover}

Kimina-Prover\cite{wang2025kimina} emphasizes training stability through KL-constrained RL. It initializes via few-shot fine-tuning, featuring innovations such as automated formalization methods for dataset construction, specialized tokenized reasoning formats for aligning informal and formal reasoning, and techniques like format filtering and negative gradient dropping to enhance stability.

\subsubsection{Goedel-Prover-V2}

Goedel-Prover-V2\cite{lin2025goedelv2} is centered around scaffolded data synthesis and iterative self-correction. Key components include scaffolded data synthesis generating progressively challenging tasks, verifier-guided self-correction leveraging Lean compiler feedback for iterative proof correction, model averaging to mitigate output diversity decline in later training stages by blending model checkpoints, and hybrid GRPO RL incorporating dynamic sampling\cite{yu2025dapo}, combined with supervised fine-tuning for efficient theorem proving with smaller parameter sizes.

These benchmark models cover three core technical directions in ATP—modular decomposition, constraint optimization, and feedback-loop enabled synthetic data. By comparing GDEPO, which focuses on innovative training algorithms, against these models, we can precisely isolate the performance gains attributed to our proposed dynamic sampling and advantage estimation mechanisms.

\subsection{Training Details}

For controlled comparisons among trainable variants, all shared hyperparameters are kept identical. The GDEPO algorithm adopts Godel-Prover-V2-8B\cite{lin2025goedelv2} as the base model and is enhanced within the Open-R1 algorithmic framework\cite{openr1}.

The learning rate is set to $1 \times 10^{-6}$, with a training batch size of 32 and an evaluation batch size of 128. Training is distributed using DeepSpeed ZeRO-3, with \texttt{bfloat16} precision and FlashAttention-2 enabled. Asymmetric clipping is applied to amplify updates from low-probability correct solutions, with $\epsilon_{\text{low}} = 0.2$ and $\epsilon_{\text{high}} = 0.28$\cite{yu2025dapo}. The optimizer is AdamW, and the learning rate follows a cosine schedule with a minimum learning rate, using a warmup ratio of 0.1. KL divergence constraints are removed during training to enhance exploration capability\cite{lin2025goedelv2}.

The number of training epochs is set to 2. The sign of the advantage function is determined solely by the correctness of the proof as judged by the Lean4 server\cite{santos2025kimina}. The reward function assigns a $+1$ reward for a correct theorem-proving trajectory and a $-1$ penalty for an incorrect one\cite{zhang2025grpolead}, with this component weighted by 1. Additionally, two auxiliary penalties are applied: a repetition penalty in the range $(-1, 1)$ based on N-gram overlap (with $N=5$), which yields positive rewards when no repetition occurs and negative rewards for severe redundancy; and a smooth length penalty also in $(-1, 1)$, computed via a cosine function of the generated sequence length—short outputs receive rewards close to $+1$, while long outputs receive rewards approaching $-1$, ensuring a smooth transition across lengths. All three reward components share equal weights to encourage concise yet correct proofs while eliminating potential bias from unequal weighting. For all baseline methods, the sampling procedure, weighted reward computation, and advantage estimation follow standard practices without modification. In contrast, our proposed GDEPO family of models performs additional sampling only when encountering ineffective training data and applies extra optimization to promote effective exploration. Outside such exploratory scenarios, GDEPO models operate identically to the baselines, adopting the same sampling strategy and reward weighting scheme.

The experimental results requiring training are reported as the average of 10 independent runs, with a different random seed used in each run to ensure the reliability of the results.

\section{Experiments}

To comprehensively evaluate the effectiveness of our algorithm from multiple perspectives, our experiments comprised four components: (a) assessing the overall performance of our method; (b) evaluating its capability to leverage challenging training data; (c) examining the efficacy of the equal-right advantage module in utilizing composite rewards; and (d) analyzing the individual contribution of each module within the GDEPO framework.

\subsection{Overall Performance Evaluation}

To evaluate the overall performance of GDEPO, we trained GOEDEL-Prover-V2 on a dataset of 30,000 samples containing 30\% challenging instances and selected the best-performing model as the representative of our method. Table\ref{tab:comp_perf} reports the performance of this model alongside SOTA baselines, including DeepSeek-Prover-V2\cite{ren2025deepseekproverv2}, Kimina-Prover\cite{wang2025kimina}, and Goedel-Prover-V2\cite{lin2025goedelv2}, along with their variants of different parameter scales. To ensure that the performance improvement of GDEPO is not affected by uncertainties, the performance data of untrained SOTA models are obtained from their respective official teams. The results show that GDEPO consistently improves performance across all three benchmarks, with an 84\% relative increase in the number of solved PutnamBench problems. This demonstrates that GDEPO can effectively unlock additional performance potential even in models that have approached their performance ceilings, highlighting its strong capability in learning from challenging samples. State-of-the-art models struggle to leverage additional computation for further performance gains, whereas our method effectively introduces meaningful computation.

\begin{table}[htbp]
\centering
\caption{Performance Comparison of GDEPO-Prover with Existing SOTA Models and Their Variants Across Model Sizes.}
\label{tab:comp_perf}
\resizebox{\linewidth}{!}{ 
\begin{tabular}{@{}cccc@{}}  
\toprule  
\multirow{2}{*}{Model}      & MiniF2F-test   & MathOlympiadBench & PutnamBench \\ 
\cmidrule(lr){2-4}  
                            & Pass Rate (\%)  & \multicolumn{2}{c}{Number of Problem Solved}         \\ 
\midrule  
DeepSeek-Prover-V2-7B       & 75.6          & 7                 & 9           \\
DeepSeek-Prover-V2-671B     & 82.4          & 50                & 22          \\
Kimina-Prover-8B            & 78.3          & 26                & 15          \\
Kimina-Prover-72B           & 84.0          & 58                & 24          \\
Goedel-Prover-V2-8B         & 84.6          & 60                & 25          \\
\textbf{GDEPO-Prover-8B(Ours)} & \textbf{86.2} & \textbf{65}       & \textbf{46}  \\ 
\bottomrule  
\end{tabular}
}
\end{table}

\subsection{Utilization Effect of Challenging Training Data Evaluation}

\begin{figure*}[!t]
\centering
\subfloat[]{\includegraphics[width=2.5in]{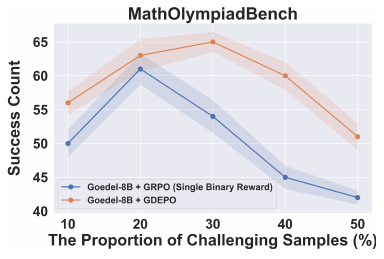}%
\label{challenging_samples_matholympiad}}
\hfil
\subfloat[]{\includegraphics[width=2.5in]{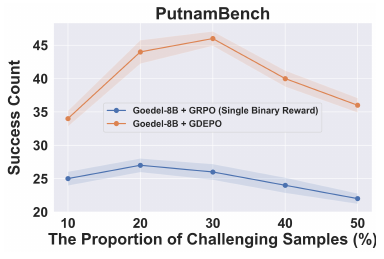}%
\label{challenging_samples_putnam}}
\caption{Performance comparison of the same base model trained with GRPO using a single binary reward versus GDEPO employing composite rewards, under varying proportions of challenging samples: (a) performances on MathOlympiadBench; (b) performances on PutnamBench.}
\label{fig:challenging_samples}
\end{figure*}

To evaluate how effectively our method leverages challenging samples in the training data, we fixed the total number of training problems at 30,000 and constructed five datasets with varying proportions of challenging samples: 10\%, 20\%, 30\%, 40\%, and 50\%. Fig.\ref{challenging_samples_matholympiad}-\ref{challenging_samples_putnam} compare the performance of the same base model trained using the standard GRPO method with a single binary reward against that trained with GDEPO employing a composite reward, across these different ratios of challenging data. The results show that GDEPO consistently outperforms GRPO, with the performance gap widening as the proportion of challenging samples increases. This indicates that GDEPO makes more effective use of challenging samples that the model initially fails to solve within a batch. Moreover, by performing additional sampling on such challenging instances that do not yield correct outputs on the first attempt, GDEPO effectively increases the equivalent batch size, thereby enhancing the probability of generating low-likelihood correct solutions and rapidly amplifying their output probabilities through the equal-right advantage.

\subsection{Effectiveness of Equal-Right Advantage Evaluation}

To assess the contribution of the equal-right advantage module to the effective utilization of composite rewards, we compare three variants: GRPO with a single binary reward, GRPO with composite rewards, and GRPO equipped with the proposed equal-right advantage. As shown in Fig.~\ref{fig:challenging_samples}, each algorithm is trained using the proportion of challenging samples that yields its best performance. Specifically, GRPO equipped with the proposed equal-right advantage is trained on 30{,}000 samples containing 30\% challenging samples, whereas GRPO under both reward configurations is trained on 30{,}000 samples containing 20\% challenging samples. The composite reward function used in GRPO matches GDEPO exactly in both reward types and weights to eliminate any confounding effects from reward weighting. As shown in Table\ref{tab:algorithm_perf}, GDEPO reduces both the average output length and repetition rate without compromising success rate, by properly computing the advantage based on composite rewards. In the table, ``number of tokens'' denotes the average token count across all completions, and ``similarity rate'' refers to the average 5-gram repetition rate. While directly applying composite rewards in GRPO also suppresses output length and redundancy, it incurs a noticeable drop in correctness—a critical metric. In contrast, GDEPO preserves correctness by using it to determine the sign of the advantage function, thereby ensuring a proper policy optimization direction, while leveraging auxiliary rewards to modulate the magnitude of policy updates and balance multiple objectives.

\begin{table*}[htbp]
\centering
\caption{Performance Comparison of the Equal-Right Advantage Module Against GRPO with Single Binary Reward and GRPO with Composite Reward Functions.}
\label{tab:algorithm_perf}
\resizebox{\linewidth}{!}{
\begin{tabular}{@{}cccccccccc@{}} 
\toprule  
\multirow{2}{*}{Algorithm} & \multicolumn{3}{c}{MiniF2F-test} & \multicolumn{3}{c}{MathOlympiad Bench} & \multicolumn{3}{c}{Putnam Bench} \\ 
\cmidrule(lr){2-4} \cmidrule(lr){5-7} \cmidrule(lr){8-10}
& Pass Rate ($\%$) $\uparrow$ & Tokens $\downarrow$ & Simi. Rate ($\%$) $\downarrow$ & Prob. Solved $\uparrow$ & Tokens $\downarrow$ & Simi. Rate ($\%$) $\downarrow$ & Prob. Solved $\uparrow$ & Tokens $\downarrow$ & Simi. Rate ($\%$) $\downarrow$ \\ 
\midrule  
Goedel-Prover-V2-8B with GRPO (Single Binary Reward) & 84.8 & 9640.89 & 49.29 & 61 & 28692.96 & 81.84 & 27 & 30148.48 & 85.23 \\
Goedel-Prover-V2-8B with GRPO & 83.2 & 7903.04 & 46.99 & 59 & 14906.10 & 75.59 & 24 & 26215.64 & 80.76 \\
$\delta$ Perf. & -1.6 & -1737.85 & -2.30 & -2 & -13786.86 & -6.25 & -3 & -3932.84 & -4.47 \\
\textbf{Goedel-Prover-V2-8B with Equal-Right Advantage} & \textbf{85.3} & \textbf{9008.35} & \textbf{48.22} & \textbf{62} & \textbf{15632.41} & \textbf{77.33} & \textbf{31} & \textbf{28510.36} & \textbf{83.87} \\
$\delta$ Perf. & +0.5 & -632.54 & -1.07 & +1 & -13060.55 & -4.51 & +4 & -1638.12 & -1.36 \\ 
\bottomrule  
\end{tabular}
}

\smallskip 
\raggedright
The performance difference relative to Goedel-Prover-V2-8B with GRPO (Single Binary Reward) is indicated as "$\delta$ Perf.". The number of tokens is indicated as "Tokens". The similarity rate is indicated as "Simi. Rate". The number of problem solved is indicated as Prob. Solved. Higher values are preferred for metrics marked with $\uparrow$, and lower values are preferred for those marked with $\downarrow$.
\end{table*}

\subsection{Ablation Evaluation}

To evaluate the individual contributions of dynamic additional sampling, dynamic additional iterations, and the equal-right advantage module in GDEPO, we conduct an ablation study. All experiments are trained on a fixed dataset of 30,000 problems containing 30\% challenging samples. Table\ref{tab:Ablation_study} presents the ablation results across three evaluation benchmarks. Since dynamic additional iterations depend on the outcomes of dynamic additional sampling, we do not test the iteration module in isolation. The results show that each component contributes to performance gains to varying degrees. Collectively, the ablation study confirms the importance of every module in the GDEPO framework. Dynamic additional sampling consistently improves model performance, with more pronounced gains on harder datasets, indicating that insufficient exploration is one of the major bottlenecks for challenging problems. Dynamic additional iterations further amplify this benefit by exploiting successful trajectories discovered through additional sampling. Meanwhile, as shown in Table~\ref{tab:algorithm_perf}, the equal-right advantage provides a complementary gain by preserving verifier-consistent gradient updates under the human-preference constraint, without requiring additional sampling. Together, these results reveal a clear division of roles: dynamic sampling enhances exploration, dynamic iterations strengthen the learning from newly discovered solutions, and equal-right advantage ensures reliable policy updates under composite rewards.

\begin{table}[htbp]
\centering
\caption{Ablation study results, showing the impact of each component on overall performance.}
\label{tab:Ablation_study}
\resizebox{\linewidth}{!}{
\begin{tabular}{@{}cccccc@{}} 
\toprule  
\multicolumn{3}{c}{Module} & MiniF2F-test & MathOlympiadBench & PutnamBench \\ 
\cmidrule(lr){1-3} \cmidrule(lr){4-6}
D.A.S. & E.R.A. & D.A.I. & Pass Rate (\%) & \multicolumn{2}{c}{Number of Problems Solved} \\ 
\midrule  
-- & -- & -- & 84.6 & 60 & 25 \\
\checkmark & -- & -- & 84.9 (+0.3) & 63 (+3) & 36 (+11) \\
-- & \checkmark & -- & 85.3 (+0.7) & 62 (+2) & 31 (+6) \\
\checkmark & -- & \checkmark & 85.8 (+1.2) & 64 (+4) & 41 (+16) \\
\textbf{\checkmark} & \textbf{\checkmark} & \textbf{\checkmark} & \textbf{86.2 (+1.6)} & \textbf{65 (+5)} & \textbf{46 (+21)} \\ 
\bottomrule  
\end{tabular}
}

\smallskip 
\raggedright 
The baseline is Godel-Prover-V2-8B. The dynamic additional sampling is indicated as ``D.A.S.''. The equal-right advantage is indicated as ``E.R.A.''. The dynamic additional iterations is indicated as ``D.A.I.''.
\end{table}

\section{The Importance of Advantage}

This section provides a supplementary analysis of the GRPO algorithm\cite{shao2024grpo}, showing that the sign of the advantage solely determines the direction of change in the trajectory's output probability. Furthermore, when a correctness-based reward is combined with other components in a composite reward function, it is mathematically possible for a correct trajectory to receive a negative advantage due to penalties from the auxiliary rewards.

\subsection{Advantage Sign as Determinant of Trajectory Probability Direction}
The gradient of the objective function after removing the KL divergence is given by
\begin{equation}
\label{eq:grpo_J_final_compact}
\nabla_\theta \mathcal{J}(\theta) = \mathbb{E} \left[ \frac{1}{G T_i} \sum_{\substack{i=1 \\ t=1}}^{G,\,T_i} \min \bigl[ \nabla_\theta \rho_{i,t}^{(s)} \hat{A}_{i,t}^*,\, \nabla_\theta \bar{\rho}_{i,t}^{(s)} \hat{A}_{i,t}^* \bigr] \right]
\end{equation}
where the auxiliary quantities are defined as
\begin{subequations}
\label{eq:grpo_aux_J_final_compact}
\begin{align}
\rho_{i,t}^{(s)} &= \frac{\pi_{\theta^{(s)}}(o_{i,t}^* \mid q, o_{i,<t}^*)}{\pi_{\theta_{\text{old}}^{(s)}}(o_{i,t}^* \mid q, o_{i,<t}^*)}, \\
\bar{\rho}_{i,t}^{(s)} &= \mathrm{clip}\bigl(\rho_{i,t}^{(s)},\, 1 - \varepsilon_{\text{low}},\, 1 + \varepsilon_{\text{high}}\bigr).
\end{align}
\end{subequations}
Here, $\pi_{\theta_{\text{old}}^{(s)}}(o_{i,t}^* \mid q, o_{i,<t}^*)$ denotes the probability under the old policy and remains constant with respect to $\theta$. The expectation operator $\mathbb{E}$ is representing expectation over the question distribution and the effective response groups sampled from the old policy $\mathbb{E}_{\substack{q \sim P(Q) \\ \mathcal{G}_{q^*} \sim \pi_{\theta_{\text{old}}^{(s)}}}}$. Consequently, it can be treated as a fixed quantity during differentiation. The gradients $\nabla_\theta \rho_{i,t}^{(s)}$ and $\nabla_\theta \bar{\rho}_{i,t}^{(s)}$ can be further decomposed:
\begin{subequations}
\label{eq:grpo_aux_J_final_compact_2}
\begin{align}
\nabla_\theta\rho_{i,t}^{(s)} &= \frac{\nabla_\theta\pi_{\theta^{(s)}}(o_{i,t}^* \mid q, o_{i,<t}^*)}{\pi_{\theta_{\text{old}}^{(s)}}(o_{i,t}^* \mid q, o_{i,<t}^*)}, \\
\nabla_\theta\bar{\rho}_{i,t}^{(s)} &= \nabla_\theta\mathrm{clip}\bigl(\rho_{i,t}^{(s)},\, 1 - \varepsilon_{\text{low}},\, 1 + \varepsilon_{\text{high}}\bigr).
\end{align}
\end{subequations}

\begin{figure}
\centering
\includegraphics[width=0.95\linewidth]{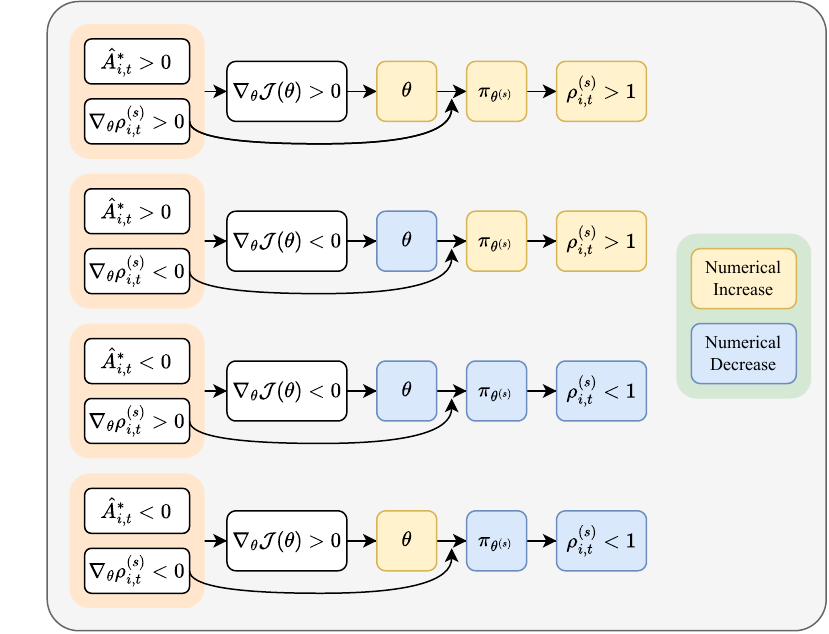}
\caption{Relationship between the sign of advantage and the direction of change in trajectory output probabilities.}
\label{fig:advantage_sign}
\end{figure}

At the first iteration, $\rho_{i,t}^{(s)} = 1$ and $\nabla_\theta \bar{\rho}_{i,t}^{(s)} = \nabla_\theta \rho_{i,t}^{(s)}$, rendering the clipping operation inconsequential. The gradient $\nabla_\theta \rho_{i,t}^{(s)}$ fundamentally reflects the alignment between the direction of change in the policy probability and the direction of parameter update $\theta$. Employing gradient ascent to maximize the objective function, the relationship between the sign of the advantage and the direction of change in the trajectory probability during the first iteration is illustrated in Fig.\ref{fig:advantage_sign}. Moreover, it follows that when $\rho_{i,t}^{(s)} \neq 1$, its relative magnitude with respect to unity is uniquely determined by the sign of the advantage. Specifically, whenever $\rho_{i,t}^{(s)}$ exceeds the upper clipping bound, $\hat{A}_{i,t}^* > 0$ holds invariably, ensuring that the minimum operator always selects $\bar{\rho}_{i,t}^{(s)}$; the converse also holds symmetrically. Consequently, the objective function—after removal of the KL divergence term—and its gradient can be simplified.

\begin{subequations}
\label{eq:grpo_J_final_compact_2}
\begin{align}
\mathcal{J}(\theta) &= \mathbb{E}_{\substack{q \sim P(Q) \\ \mathcal{G}_{q^*} \sim \pi_{\theta_{\text{old}}^{(s)}}}} \left[ \frac{1}{G T_i} \sum_{\substack{i=1 \\ t=1}}^{G,\,T_i} \hat\rho_{i,t}^{(s)} \hat{A}_{i,t}^* \right], \\
\nabla_\theta \mathcal{J}(\theta) &= \mathbb{E}_{\substack{q \sim P(Q) \\ \mathcal{G}_{q^*} \sim \pi_{\theta_{\text{old}}^{(s)}}}} \left[ \frac{1}{G T_i} \sum_{\substack{i=1 \\ t=1}}^{G,\,T_i} \nabla_\theta \hat\rho_{i,t}^{(s)} \hat{A}_{i,t}^* \right]
\end{align}
\end{subequations}
where the auxiliary quantities are defined as
\begin{subequations}
    \label{eq:grpo_J_final_compact3}
    \begin{align}
    \hat{\rho}_{i,t}^{(s)} &= 
    \begin{cases}
        \rho_{i,t}^{(s)}, & \rho_{i,t}^{(s)} \in \bigl(1 - \varepsilon_{\text{low}},\, 1 + \varepsilon_{\text{high}}\bigr) \\
        \bar{\rho}_{i,t}^{(s)}, & \rho_{i,t}^{(s)} \notin \bigl(1 - \varepsilon_{\text{low}},\, 1 + \varepsilon_{\text{high}}\bigr)
    \end{cases}, \\
    \nabla_\theta \hat{\rho}_{i,t}^{(s)} &= 
    \begin{cases}
    \nabla_\theta\rho_{i,t}^{(s)}, & \rho_{i,t}^{(s)} \in \bigl(1 - \varepsilon_{\text{low}},\, 1 + \varepsilon_{\text{high}}\bigr) \\
    0, & \rho_{i,t}^{(s)} \notin \bigl(1 - \varepsilon_{\text{low}},\, 1 + \varepsilon_{\text{high}}\bigr)
    \end{cases}.
    \end{align}
\end{subequations}

It is proven that the $\min$ function, in extreme cases, enforces the gradient to zero, thereby preventing excessive deviation in the output probability of the trajectory across multiple iterations. In subsequent iterations, even when $\rho_{i,t}^{(s)} \neq 1$, the sign of the advantage and the direction of change in the trajectory's output probability remain consistent with the procedure illustrated in Fig.\ref{fig:advantage_sign}.

In summary, the sign of the advantage function uniquely determines the direction of change in the trajectory output probability.

\subsection{Possibility of Negative Advantages on Correct Trajectories under Composite Rewards}

The advantage function in the GRPO algorithm is computed as follows:
\begin{equation}
\label{eq:grpo_advantage_ori}
\hat{A}_{i,t} = \tilde{r}_i = \frac{r_i - \text{mean}(r)}{\text{std}(r)}
\end{equation}

where $r_i$ denotes the total reward of the $i$-th output. $\mathrm{mean}(r)$ and $\mathrm{std}(r)$ represent the mean and standard deviation of the rewards within the group, respectively. The advantage function for all tokens corresponding to a given output is identical and equal to the normalized total reward of that output (i.e., $\hat{A}_{i,t}$ remains constant across all timesteps $t$ for the same output). 

For a query $q$, a sampled trajectory is denoted as $\mathcal{G}_q = \{ o_1, o_2, \ldots, o_G \}$. The correctness binary reward is represented by $\mathcal{L}_q = \{ l_1, l_2, \ldots, l_G \}$, where each $l_i \in \{0, 1\}$, with $1$ indicating a correct output and $0$ an incorrect one. The auxiliary reward is denoted as $\mathcal{R}_q = \{ \mathbf{r}_1, \mathbf{r}_2, \ldots, \mathbf{r}_G \}$, where each $\mathbf{r}_i = ( r_{i,1}, r_{i,2}, \ldots, r_{i,K} )$ is a $K$-dimensional auxiliary reward vector whose components satisfy $r_{i,j} \in [0, 1]$ for all $j = 1, \ldots, K$. 

The weight assigned to the correctness binary reward is $\alpha > 0$, and the weights for the auxiliary rewards are given by $w = (w_1, \ldots, w_K)$, subject to the constraint $\sum_{j=1}^K w_j = 1$. The raw weighted reward for output $o_i^{(k)}$ is defined as $s_i^{(k)} = \alpha \cdot l_i^{(k)} + \sum_{j=1}^K w_j \cdot r_{i,j}^{(k)}$, with its empirical mean and standard deviation given respectively by $\mu_s^{(k)} = \frac{1}{G} \sum_{i=1}^G s_i^{(k)}$ and $\sigma_s^{(k)} = \sqrt{\frac{1}{G-1} \sum_{i=1}^G (s_i^{(k)} - \mu_s^{(k)})^2}$. Eq.\ref{eq:grpo_advantage_ori} can thus be reformulated as follows:

\begin{equation}
\label{eq:grpo_advantage_1}
\hat{A}_i = \frac{s_i - \mu_s}{\sigma_s}
\end{equation}
The sign of the advantage is determined by the numerator: $\hat{A}_i < 0 \iff s_i < \mu_s$.

To simplify the analysis while maintaining generality, the following reasonable assumptions are introduced. 1) \textit{Reward Independence Assumption}: The correctness indicator $l_i$ and the auxiliary reward vector $\mathbf{r}_i$ are mutually independent, i.e., $\mathbb{E}[l_i r_{i,j}] = \mathbb{E}[l_i] \mathbb{E}[r_{i,j}]$ for all $j$. 2) \textit{Non-degeneracy Within Group Assumption}: The expected correctness rate satisfies $p = \mathbb{E}[l_i] \in (0,1)$, and the expected auxiliary rewards satisfy $\mu_j = \mathbb{E}[r_{i,j}] \in (0,1)$ for all $j$, ensuring that both correct and incorrect outputs exist within the group. Moreover, the standard deviation of the total reward within the group satisfies $\sigma_s > 0$, thereby excluding the degenerate case where all outputs receive identical rewards.3) \textit{Non-zero Weight Assumption}: The weight assigned to the correctness reward satisfies $\alpha > 0$, and each auxiliary reward weight satisfies $w_j > 0$, guaranteeing that both reward components contribute non-trivially to the total reward.

The outputs within the group are partitioned into two subsets based on correctness: the set of correct outputs $\mathcal{G}_1 = \{ o_i \mid l_i = 1 \}$, with cardinality $G_1 = pG$ and weighted average auxiliary reward $\bar{r}_1 = \sum_{j=1}^K w_j \cdot \frac{1}{G_1} \sum_{o_i \in \mathcal{G}_1} r_{i,j}$; and the set of incorrect outputs $\mathcal{G}_0 = \{ o_i \mid l_i = 0 \}$, with cardinality $G_0 = (1-p)G$ and weighted average auxiliary reward $\bar{r}_0 = \sum_{j=1}^K w_j \cdot \frac{1}{G_0} \sum_{o_i \in \mathcal{G}_0} r_{i,j}$.

Consequently, the group-level mean of the total reward can be decomposed as follows:
\begin{equation}
\label{eq:grpo_weighted_reward}
\mu_s = \alpha p + p \bar{r}_1 + (1-p) \bar{r}_0
\end{equation}
The average total reward of the correct subset is given by:
\begin{equation}
\label{eq:grpo_correct_weighted_reward}
\mu_{s,1} = \alpha + \bar{r}_1
\end{equation}
There exists a correct output $o_i \in \mathcal{G}_1$ such that $\hat{A}_i < 0$ if:
\begin{equation}
\label{eq:grpo_general_condition}
\mu_{s,1} < \mu_s
\end{equation}
The necessity is proved as follows. Suppose there exists $o_i \in \mathcal{G}_1$ for which $\hat{A}_i < 0$. Then, by definition of the advantage function, $s_i < \mu_s$. Since $\mu_{s,1} = \frac{1}{G_1} \sum_{o_i \in \mathcal{G}_1} s_i$ denotes the average total reward over all correct outputs, if $s_i \geq \mu_s$ held for every $o_i \in \mathcal{G}_1$, it would follow that $\mu_{s,1} \geq \mu_s$, leading to a contradiction. Hence, it must be that $\mu_{s,1} < \mu_s$.

For sufficiency, suppose $ \mu_{s,1} < \mu_s $. By the definition of the arithmetic mean, there necessarily exists at least one $ o_i \in \mathcal{G}_1 $ such that $ s_i < \mu_s $. Given that $ \sigma_s > 0 $—which follows from the non-degeneracy of rewards within the group (as identical rewards would render the advantage function undefined)—it follows that $ \hat{A}_i < 0 $. Consequently, there exists a correct output exhibiting a negative advantage.

In summary, when a correctness-based reward is incorporated into a composite reward structure, it is mathematically possible for a correct trajectory to yield a negative advantage due to the influence of other reward components.
\section{Conclusion}

This paper addresses critical limitations in current RL based ATP methods concerning training efficiency and optimization direction by proposing the GDEPO framework. GDEPO systematically enhances the learning efficacy of LLMs in formal reasoning through three key mechanisms. First, the Lean verifier is used to precisely identify and discard entirely ineffective training samples, while a supplementary sampling strategy actively explores potentially correct proof paths. This introduces effective additional computation into models whose rewards have already converged, thereby improving performance. Second, we design the key equal-right advantage mechanism, which preserves the verifier's binary correctness signal as the optimization direction and fuses it with multi-dimensional auxiliary rewards to enhance the expressiveness of the policy gradient. This yields additional valid and correct computation for the model without requiring extra sampling or additional training data. Third, it applies dynamic-iteration backpropagation on challenging samples that initially fail but eventually succeed, accelerating the model's mastery of difficult yet solvable problems.

We conduct comprehensive experiments on three progressively challenging formal mathematical benchmarks: MinF2F-test, MathOlympiadBench, and PutnamBench. Results demonstrate that GDEPO significantly outperforms existing GRPO-based training paradigms, achieving substantial gains in both proof success rate and data utilization efficiency. Ablation studies further confirm that all three components are essential and act synergistically.

This work not only delivers a more efficient and robust training framework for LLM-driven ATP systems but also offers a novel perspective on designing RL algorithms for tasks equipped with precise verification mechanisms. With the advancement of AI, nearly all collectible training data have already been exhausted, and the difficulty of acquiring or generating high-value training data continues to rise. While prior research has largely focused on increasing data quantity and quality and on crafting sophisticated reward functions, we argue that future efforts should place equal emphasis on maximizing the utility of existing data and on developing domain-informed strategies for computing trajectory advantages in RL applications.

\bibliographystyle{cas-model2-names}
\bibliography{reference.bib}

\vfill

\end{document}